%% file: arxiv.tex
\documentclass[10pt,journal,compsoc]{IEEEtran}
\ifCLASSOPTIONcompsoc
  \usepackage[nocompress]{cite}
\else
  \usepackage{cite}
\fi
\usepackage{hyperref}
\usepackage{amsmath,amsfonts}
\usepackage{array}
\usepackage{subfigure}
\usepackage{textcomp}
\usepackage{stfloats}
\usepackage{url}
\usepackage{verbatim}
\usepackage{graphicx}
\usepackage{epsfig}
\usepackage{amssymb}
\usepackage{multirow}
\usepackage{makecell} 
\usepackage{pifont}
\newcommand{\cmark}{\ding{51}}%
\newcommand{\xmark}{\ding{55}}%
\usepackage[ruled,linesnumbered]{algorithm2e}
\usepackage{algpseudocode}
\usepackage{verbatim}
\usepackage{tabularx}
\usepackage{xcolor,colortbl}
\definecolor{Gray}{gray}{0.9}
\definecolor{LightCyan}{rgb}{0.88,1,1}
\newcolumntype{C}[1]{>{\centering\arraybackslash}p{#1}}
\usepackage{balance}
\begin{document}
\title{DVMNet++: Rethinking Relative Pose Estimation for Unseen Objects}

\author{Chen Zhao, Tong Zhang, Zheng Dang, and Mathieu Salzmann
\IEEEcompsocitemizethanks{
	\IEEEcompsocthanksitem  Chen Zhao and Zheng Dang are with the Computer Vision Laboratory, École Polytechnique Fédérale de Lausanne, CH-1015 Lausanne, Switzerland. \protect\\
	E-mail: chen.zhao@epfl.ch, zheng.dang@epfl.ch.
	\IEEEcompsocthanksitem  Tong Zhang is with the Image and Visual Representation Laboratory, École Polytechnique Fédérale de Lausanne, CH-1015 Lausanne, Switzerland. \protect\\
	E-mail: tong.zhang@epfl.ch.
	\IEEEcompsocthanksitem  Mathieu Salzmann is with the Computer Vision Laboratory, École Polytechnique Fédérale de Lausanne, CH-1015 Lausanne, Switzerland, and also with
Clearspace, 1020 Renens, Switzerland. \protect\\
	E-mail: mathieu.salzmann@epfl.ch.}
} 

\markboth{}%
{Shell \MakeLowercase{\textit{et al.}}: Bare Demo of IEEEtran.cls for Computer Society Journals}

\input{sec/0_abstract}    
\maketitle

\input{sec/1_intro}
\input{sec/2_related_work}
\input{sec/3_method}
\input{sec/4_experiments}

\input{sec/5_conclusion}

\bibliographystyle{IEEEtran}
\bibliography{main}
\input{sec/biography}


    
    

\end{document}

%% file: sec/0_abstract.tex
\IEEEtitleabstractindextext{\begin{abstract}
Determining the relative pose of a previously unseen object between two images is pivotal to the success of generalizable object pose estimation. Existing approaches typically predict 3D translation utilizing the ground-truth object bounding box and approximate 3D rotation with a large number of discrete hypotheses. This strategy makes unrealistic assumptions about the availability of ground truth and incurs a computationally expensive process of scoring each hypothesis at test time. By contrast, we rethink the problem of relative pose estimation for unseen objects by presenting a Deep Voxel Matching Network (DVMNet++). Our method computes the relative object pose in a single pass, eliminating the need for ground-truth object bounding boxes and rotation hypotheses. We achieve open-set object detection by leveraging image feature embedding and natural language understanding as reference. The detection result is then employed to approximate the translation parameters and crop the object from the query image. For rotation estimation, we map the two RGB images, i.e., reference and cropped query, to their respective voxelized 3D representations. The resulting voxels are passed through a rotation estimation module, which aligns the voxels and computes the rotation in an end-to-end fashion by solving a least-squares problem. To enhance robustness, we introduce a weighted closest voxel algorithm capable of mitigating the impact of noisy voxels. We conduct extensive experiments on the CO3D, Objaverse, LINEMOD, and LINEMOD-O datasets, demonstrating that our approach delivers more accurate relative pose estimates for novel objects at a lower computational cost compared to state-of-the-art methods. Our code is released at:\href{https://github.com/sailor-z/DVMNet/}{https://github.com/sailor-z/DVMNet/}.
\end{abstract}
\begin{IEEEkeywords}
    Object pose estimation, unseen objects, two-view geometry, 3D computer vision.
\end{IEEEkeywords}
}

%% file: sec/1_intro.tex
\section{Introduction}
\label{sec:intro}

\IEEEPARstart{O}bject pose estimation plays a crucial role in 3D computer vision and robotics tasks~\cite{azuma1997survey,geiger2012we,marchand2015pose,tremblay2018deep}, aiming to produce 3D translation and 3D rotation of an object depicted in an RGB image. The vast majority of existing methods work under the assumption that the training and testing data include the same object instances, thereby limiting their applicability to scenarios that involve previously unseen objects. Recently, generalizable object pose estimation~\cite{liu2022gen6d,shugurov2022osop,sun2022onepose,wen2024foundationpose} has received growing attention, showcasing the potential to generalize to unseen objects from new categories without retraining the network. In pursuit of this generalization capability, existing methods leverage densely sampled images depicting unseen objects in diverse poses, serving as references. Object pose estimation is then carried out through template matching~\cite{liu2022gen6d,shugurov2022osop,zhao2022fusing,wen2024foundationpose} or by establishing 2D-3D correspondences~\cite{sun2022onepose,he2022onepose++,lee2024mfos}. Unfortunately, the effectiveness of these methods strongly depends on the references densely covering the viewpoints of the unseen objects, making them inapplicable to practical scenarios where only sparse reference views are available.

\begin{figure}[!t]
  \centering
  \includegraphics[width=0.9\linewidth]{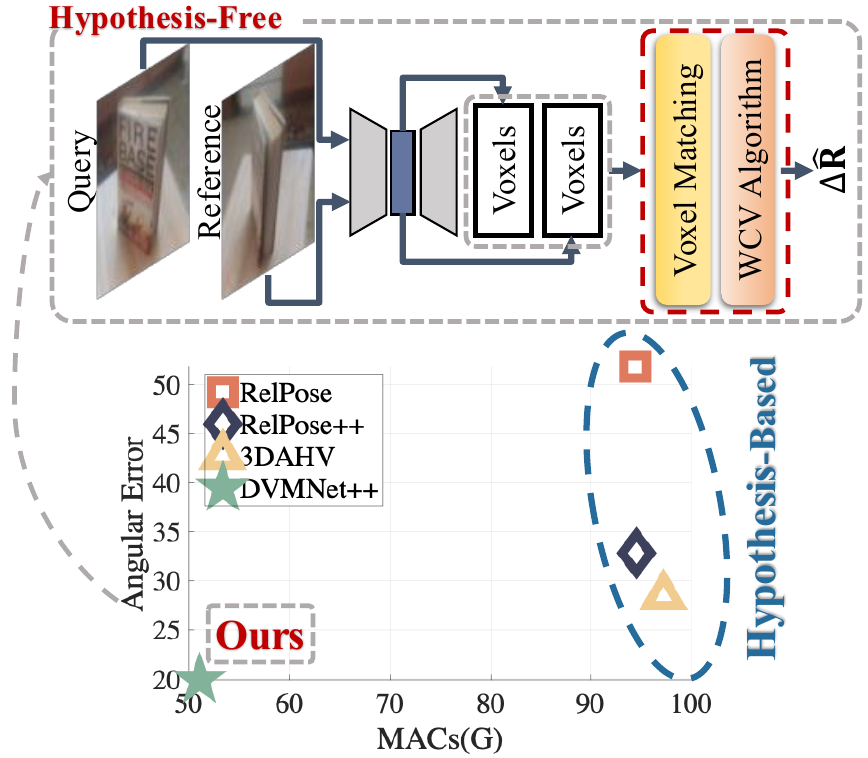}
  \caption{\textbf{Advantages of our DVMNet++ compared to hypothesis-based methods.} Hypothesis-based techniques approximate the relative object rotation by scoring numerous rotation hypotheses, leading to a high computational cost. By contrast, our DVMNet++ computes the rotation in a hypothesis-free fashion by robustly matching voxelized 3D representations of the reference and query images via a Weighted Closest Voxel algorithm. Our method strikes a favorable balance between computational cost and accuracy in relative object pose estimation, as measured by multiply-accumulate operations (MACs) and angular error.}
  \label{fig:intro}
\end{figure}

In this context, a few methods~\cite{zhang2022relpose,lin2023relpose++,zhao20233d} highlight the importance of relative object pose estimation. Unlike previous approaches in generalizable object pose estimation, these methods focus on estimating the relative pose of an unseen object between two images, i.e., a query image and a single reference image of the object. In this paper, we also work in this setting, motivated by the practical ease of obtaining a single reference image for a new object. One plausible solution is to compute the relative pose based on 2D-2D correspondences~\cite{hartley2003multiple}. However, the single-reference scenario tends to yield a significant viewpoint gap between the reference and the query. Existing studies~\cite{lin2023relpose++,zhao20233d} have shown that image-matching techniques~\cite{sarlin2020superglue,sun2021loftr} are sensitive to such pose differences. To handle this issue, the prior methods~\cite{zhang2022relpose,lin2023relpose++,zhao20233d} follow an alternative strategy of scoring multiple rotation hypotheses for the input reference-query pair, and predicting the rotation as the hypothesis with the highest score. However, this alternative comes with the drawback of requiring numerous rotation hypotheses to achieve reasonable accuracy, e.g., 500,000 in~\cite{lin2023relpose++}, which thus induces a computational burden. Moreover, we empirically found that these approaches occasionally produce unnaturally large errors. One plausible explanation is their failure to model the continuous nature of the object rotation space, as they primarily concentrate on learning to score discrete hypotheses.

Additionally, it is worth noting that the aforementioned approaches assume that the \emph{ground-truth} object bounding boxes are known, even at test time. Such ground-truth information facilitates the relative object pose estimation from two aspects: First, the object bounding box parameters are fed into a translation regression network~\cite{zhang2022relpose, lin2023relpose++}, which provides strong prior information about the object translation; second, the region containing the object is cropped from the query image, mitigating the impact of the background when estimating the object rotation. Unfortunately, since we focus on relative pose estimation for novel objects that are not included in the training set, detecting the previously unseen objects is non-trivial, particularly in cluttered scenes~\cite{hinterstoisser2012model}. Due to the reliance on the ground-truth object bounding boxes, existing methods become inapplicable in scenarios where high-accuracy object bounding boxes are unavailable.

To overcome these drawbacks, we present a new pipeline DVMNet++ that computes the relative pose of unseen objects efficiently without relying on ground-truth object bounding boxes. Our approach starts by detecting the object in the query image. Specifically, we draw inspiration from recent progress in open-vocabulary object detection~\cite{gu2021open,minderer2022simple,liu2023grounding,yao2023detclipv2}, which demonstrates promising detection accuracy for unseen objects. Since we have access to the reference image depicting the object, we describe the object via text prompts based on the reference image. The text prompts and the query image are taken as the input of an open-vocabulary object detection network~\cite{liu2023grounding} that produces object proposals. To identify the most reliable object bounding box from the generated proposals, we measure the similarity of the reference image and each proposal in high-dimensional feature space. The proposal closest to the reference in the feature space is selected as the detection result. We approximate the relative object translation and crop the object from the query using the identified bounding box. Subsequently, we achieve the hypothesis-free relative object rotation estimation by introducing a deep voxel matching network. We first voxelize the reference image and the cropped query image in a dedicated autoencoder. The encoder network lifts 2D image features to 3D voxels, leveraging cross-view 3D information. The decoder network reconstructs a masked object image from the voxels, encouraging the learned voxels to account for the object. We then align the query and reference voxels based on a score matrix that measures the voxel similarities. To handle unreliable voxels due to background, varying illumination, and other potential nuisances, we present a Weighted Closest Voxel (WCV) algorithm to facilitate robust rotation estimation. In this algorithm, each voxel-voxel correspondence is assigned a confidence score computed by utilizing both the 3D voxel objectness map and the 2D object mask learned by the autoencoder. The relative object rotation is computed by solving a weighted least-squares problem. Such an end-to-end learning mechanism eliminates the necessity for voxel-wise annotations and allows the network to directly learn rotation-aware features from RGB images. As illustrated in Fig.~\ref{fig:intro}, our DVMNet++ requires significantly fewer multiply-accumulate operations (MACs) while achieving smaller angular errors than its hypothesis-based competitors.

We perform comprehensive experiments on the CO3D~\cite{reizenstein2021common}, Objaverse~\cite{deitke2023objaverse}, LINEMOD~\cite{hinterstoisser2012model}, and LINEMOD-O~\cite{brachmann2014learning} datasets. Our method yields more accurate and robust relative pose estimates for previously unseen objects than the existing competitors. We also conduct ablation studies where the results demonstrate the effectiveness of the key components in our framework. In short, our contributions are threefold:
\begin{itemize}
    \item We eliminate the reliance on \emph{ground-truth} object bounding boxes in relative pose estimation for unseen objects by introducing a new open-set object detector.
    \item We tackle the problem of relative object rotation estimation in a \emph{hypothesis-free} manner by presenting a deep voxel matching network.
    \item We present a weighted closest voxel algorithm that robustly computes the relative object rotation from voxel-voxel correspondences in an end-to-end manner.
\end{itemize}

This paper extends our previous work~\cite{zhao2024dvmnet} by highlighting the importance of open-set object detection in the pipeline of relative object pose estimation. We integrate an open-set object detector with our previous DVMNet, achieving generalizable 6D relative object pose estimation without relying on ground-truth object bounding boxes. We also provide a more detailed analysis of our method, showcasing the robustness towards occlusion and the compatibility in the scenario of sparse references. 

The remainder of this paper is organized as follows. Section~\ref{sec:related} provides an overview of the related work. Section~\ref{sec:method} presents the detailed methodology of DVMNet++. Section~\ref{sec:experiments} reports the experimental results across several datasets and includes comprehensive ablation studies. Section~\ref{sec:conclusion} summarizes our contributions and outlines directions for future work.

%% file: sec/2_related_work.tex
\section{Related Work}
\label{sec:related}

\noindent\textbf{Instance-Level Object Pose Estimation.} 
The majority of previous deep learning approaches to object pose estimation~\cite{xiang2017posecnn,peng2019pvnet,wang2021gdr,su2022zebrapose,wang2019densefusion} tackle the problem at an instance level, assuming that the training and testing data depict the same object instances. Since the appearance of an object instance in different poses typically exhibits limited variations, these methods provide highly accurate object pose estimates. Nevertheless, they struggle to generalize to previously unseen objects during testing without retraining the network, as has been observed in the literature~\cite{liu2022gen6d,shugurov2022osop,sun2022onepose}. This limitation constrains their applicability in real-world scenarios that often involve diverse object instances. This problem has been remedied to a degree by category-level object pose estimation methods~\cite{wang2019normalized,chen2020learning,lin2022category}. In this scenario, the testing images comprise new object instances from specific categories already included in the training data. Although these methods have achieved promising generalization ability within the predefined object categories, they become ineffective when facing objects from entirely new categories.
\\

\noindent\textbf{Generalizable Object Pose Estimation.} To tackle the scenario of unseen objects from new categories, there has been growing interest in generalizable object pose estimation. When a textured 3D mesh is available for an unseen object, some approaches~\cite{wohlhart2015learning,zhao2022fusing,shugurov2022osop} suggest generating synthetic images as references by rendering the 3D mesh from various viewpoints. Given a query image that depicts this object, a template matching paradigm is utilized to identify the most similar reference and approximate the object pose in the query as that of the selected reference. Some methods bypass the need for 3D meshes by assuming the availability of multiple real reference images. Object pose estimation is then carried out by employing either a template matching strategy~\cite{liu2022gen6d} or a 3D object reconstruction technique~\cite{sun2022onepose,he2022onepose++,wen2024foundationpose}. Nevertheless, all of these methods rely on having access to dense-view reference images, which limits their applicability in scenarios where only sparse reference views are available.
\\

\noindent\textbf{Relative Object Pose Estimation.} In such a context, several studies~\cite{zhang2022relpose,lin2023relpose++,zhao20233d} have highlighted the importance of relative object pose estimation. These methods stand out in generalizable object pose estimation due to their key advantage of requiring only a single reference image. The objective of these methods is to estimate the relative object pose between the input query image and the reference. Since the single-reference assumption tends to result in a large object pose difference between the query and the reference, unseen object pose estimation becomes more challenging. Intuitively, one could establish pixel-pixel correspondences between the two images and compute the relative object pose based on multi-view geometry~\cite{hartley2003multiple}. However, as reported in the literature~\cite{lin2023relpose++,zhao20233d} and also in our experiments, image-matching techniques~\cite{sarlin2020superglue,sun2021loftr,goodwin2022zero} have difficulty in delivering accurate pose estimates when confronted with large object pose differences. To address this issue, existing methods~\cite{zhang2022relpose,lin2023relpose++,zhao20233d} suggest approximating the relative object rotation via a discrete set of rotation hypotheses, and learning to maximize the score of the positive hypotheses. Since object rotation lies in a continuous space~\cite{zhou2019continuity}, accurately approximating the rotation necessitates a vast number of rotation hypotheses, which makes such a hypothesis-based approach computationally expensive. Moreover, scoring discrete samples lacks an understanding of the continuous rotation distribution, leading to failure cases with unnaturally high rotation estimation errors. As an alternative, in~\cite{wang2023posediffusion,zhang2024cameras}, a diffusion mechanism is employed to regress the pose parameters. The iterative denoising process during inference nonetheless makes this approach time-consuming. By contrast, we present a hypothesis-free technique that is capable of computing the relative object rotation in a single pass via deep voxel matching.
\\

\noindent\textbf{Open-Set Object Detection.} The aforementioned relative object pose estimation approaches are designed for object-centric scenarios where the object is positioned at the center of the image and thus the object bounding box is easy to obtain. In this context, the ground-truth object bounding box is assumed to be available. It is employed to predict the relative object translation~\cite{zhang2022relpose,lin2023relpose++} and crop the object from the query image. However, in some applications, especially in cluttered scenes, effectively detecting a previously unseen object is challenging. In recent years, open-set object detection~\cite{gu2021open,minderer2022simple,liu2023grounding,yao2023detclipv2} has received significant attention due to its capacity to identify novel objects from unseen categories. Some pioneering methods have been developed, incorporating open-set object detection into object pose estimation. For instance, CNOS~\cite{nguyen2023cnos} and SAM-6D~\cite{lin2024sam} propose to utilize SAM~\cite{kirillov2023segment} to generate object proposals. The object mask is selected by matching the proposals with templates based on DINOv2~\cite{oquab2023dinov2} feature similarities. Gen6D~\cite{liu2022gen6d} and LocPoseNet~\cite{zhao2024locposenet} predict the bounding box parameters building upon a template matching mechanism. However, a notable limitation of these approaches is their reliance on dense-view reference images, making them inapplicable to our single-reference setting. Therefore, we present a new unseen object detection approach that leverages multi-modal reference information from a single view. As will be witnessed by our experiments, it yields robust detection results for previously unseen objects.

\begin{figure}[t]
	\subfigure[Input]
	{ \includegraphics[height=2.7cm]{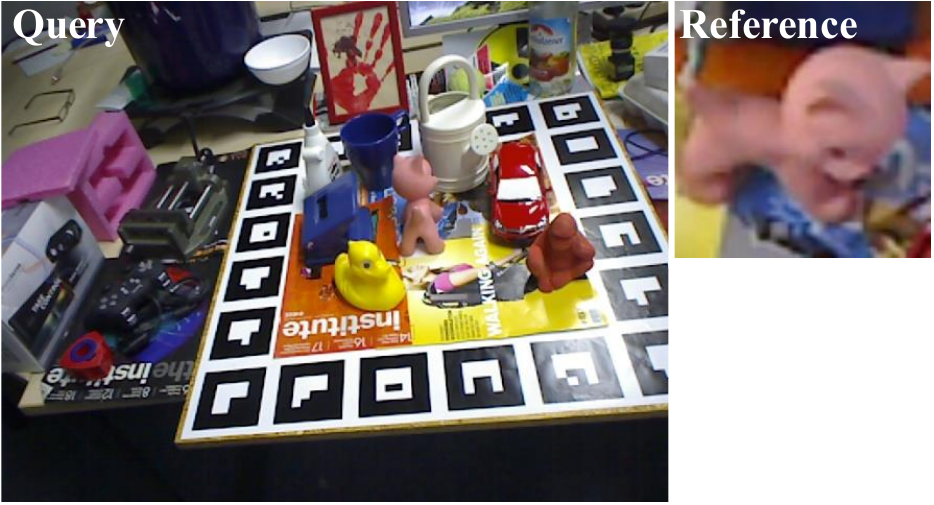}\label{formulation_a}}	
	\subfigure[Output]
	{ \includegraphics[height=2.7cm]{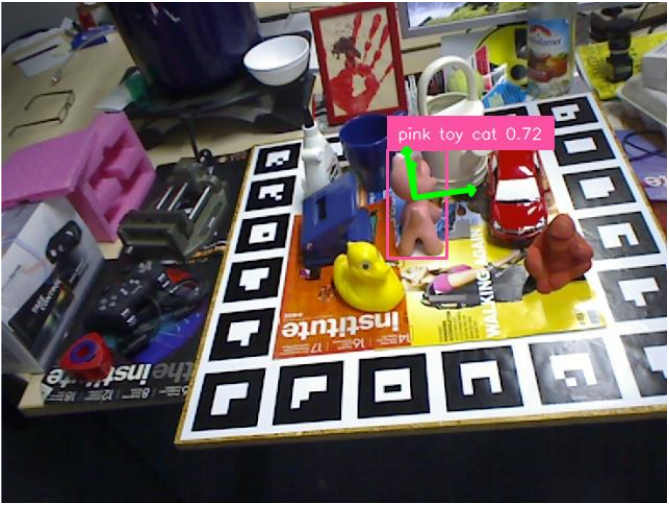}\label{formulation_b}}
	\caption{\textbf{Problem formulation.} (a) Input to our method, consisting of a query image and a reference image. (b) Our goal is to identify the corresponding object in the query image and estimate the object translation and rotation based on the reference image. We represent the predicted translation and rotation as a bounding box and green arrows, respectively.}
	\label{fig:formulation}
\end{figure}

%% file: sec/3_method.tex
\section{Method}
\label{sec:method}

\begin{figure*}[!t]
    \centering	
    \includegraphics[width=1.0\linewidth]{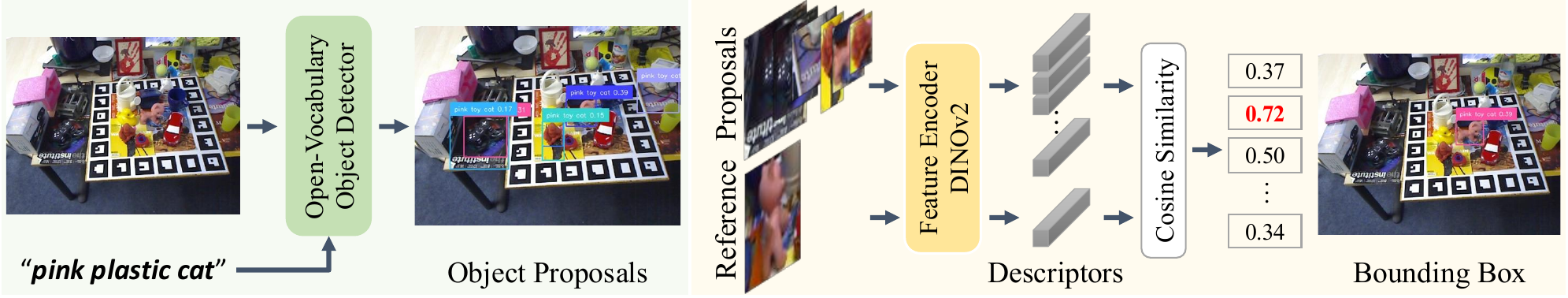}
    \caption{\textbf{Open-set object detection.} We incorporate an open-set object detection module in our relative object pose estimation framework, utilizing multi-modal reference information. Given the reference image, we describe the object appearance using text prompts. An open-vocabulary object detection network takes these prompts and the query image as input, and predicts a set of object proposals. Since the generated proposals may include outliers, we propose identifying the most reliable prediction using an image retrieval technique. We encode the reference image and proposals to feature descriptors by utilizing a pretrained DINOv2 encoder. The final detection result is determined as the proposal with the highest cosine similarity score.}
    \label{fig:detection}
\end{figure*}

\subsection{Problem Formulation}
We tackle the problem of estimating the 6D pose $\mathbf{P}$ for a previously unseen object depicted in an RGB image $\mathbf{I}_q$. The object pose consists of a 3D translation $\mathbf{T}$ and a 3D rotation $\mathbf{R}$. In this scenario, the objects present in the testing set $\Omega_{\text{test}}$ differ from those in the training set $\Omega_{\text{train}}$, and the goal is to handle the unseen objects without retraining the network. Furthermore, we assume that one RGB image $\mathbf{I}_r$ depicting the object is given as a reference, following the setting in~\cite{zhao20233d,zhao2024dvmnet}. Notably, both 3D CAD models and dense-view reference images are unavailable in this setting.

Therefore, the goal is to estimate the relative object pose $\Delta{\mathbf{P}}$ between the query image $\mathbf{I}_q$ and the reference image $\mathbf{I}_r$. The challenges of this problem lie in the ability to generalize to the unseen objects in $\Omega_{\text{test}}$ and in the need for robustness to the large object pose difference between $\mathbf{I}_q$ and $\mathbf{I}_r$. As illustrated in Fig.~\ref{fig:formulation}, unlike previous methods~\cite{lin2023relpose++,zhao20233d,zhao2024dvmnet}, we do not assume a known bounding box to identify the object in the query image. Instead, an open-set object detection approach is required to yield object bounding box parameters $(c_x, c_y, w, h)$, where $c_x$ and $c_y$ denote the center of the bounding box, and $w$ and $y$ indicate the width and height. Furthermore, to estimate the relative pose, we define the reference object coordinate system such that its origin aligns with that of the canonical coordinate system, i.e., setting $\mathbf{T}_r = {[0, 0, 0]}^{T}$. The ground-truth relative object translation and rotation are then defined as $\Delta{\mathbf{T}}=\mathbf{T}_q$ and $\Delta{\mathbf{R}}=\mathbf{R}_q\mathbf{R}_r^{T}$, respectively. According to the pinhole camera model, $\mathbf{T}_q$ can be computed as 
\begin{align}
\label{eq:translation}
\mathbf{T}_q = d_q \mathbf{K}_q {[u_q, v_q, 1]}^{T},
\end{align}
where $\mathbf{K}_q$ denotes the camera intrinsic, $d_q$ represents the depth of the object center, and $(u_q, v_q)$ indicates the 2D object center in the query image. In this paper, we assume known camera intrinsics~\cite{li2018deepim} and approximate the 2D object center as the center of the detected bounding box, i.e., $u_q=c_x$ and $v_q=c_y$. Since we do not have access to the 3D object model, the translation estimate is inherently ambiguous and can only be determined up to a scale factor. To address the scale ambiguity, we evaluate the translation estimation in terms of the angular error~\cite{wang2023posediffusion}. We will elaborate on this metric in Section~\ref{sec:experiments}.

To estimate the relative object rotation, previous hypothesis-based approaches~\cite{zhang2022relpose,lin2023relpose++,zhao20233d} approximate $\Delta{\mathbf{R}}$ by sampling discrete rotation hypotheses and maximizing the score of the positive samples. This can be formulated as 
\begin{align}
\label{eq:energy}
\Delta{\hat{\mathbf{R}}}=\underset{{\Delta{\mathbf{R}_{i}}\in \mathcal{R}}}{\arg\max} \ f(\mathbf{I}_q, \mathbf{I}_r, \Delta{\mathbf{R}}_i),
\end{align}
where $\mathcal{R}$ denotes the set of discrete rotation hypotheses.
Achieving a decent approximation accuracy requires a large number of hypotheses, e.g., 500,000  in~\cite{lin2023relpose++}. By contrast, we present a hypothesis-free technique that computes $\Delta{\mathbf{R}}$ in a single pass as $\Delta{\hat{\mathbf{R}}}=g(\mathbf{I}_q, \mathbf{I}_r)$. 

\subsection{Open-Set Object Detection}
Recall that existing object detection methods~\cite{nguyen2023cnos,lin2024sam,zhao2024locposenet} in generalizable object pose estimation rely on dense-view reference images. Consequently, we introduce a new open-set object detection approach that is applicable in our single-reference setting. We propose to facilitate the open-set object detection by leveraging multi-modal reference information, consisting of the RGB image and a natural language description.

Specifically, as shown in Fig.~\ref{fig:detection}, we describe the object in terms of its attributes and category based on the available reference image. The resulting text prompts, along with the query image, are employed as inputs to an open-vocabulary object detector~\cite{liu2023grounding}. The output of this detector consists of $M$ object proposals, which lets us formulate the detection process as 
\begin{align}
\label{eq:detection}
\{p_1, p_2,\dots,p_M\}=f_d(\mathbf{I}_q, t_q|\theta),
\end{align}
where $p_i$ denotes an object proposal, $f_d$ indicates the detection network~\cite{liu2023grounding} with pretrained parameters $\theta$, and $t_q$ represents the text prompt.
Each proposal denotes a region with parameters $(c_x^{i}, c_y^{i}, w_i, h_i, s_i)$ in the query image, which is likely to contain the object, with $s_i$ a confidence score indicating the reliability of the prediction. 

In our initial experiments, we observed the object proposals to be noisy. As illustrated Fig.~\ref{fig:detection}, some detected regions contain the wrong objects. To identify the most reliable result from the candidates, we present an image retrieval strategy utilizing the reference image. We crop the regions from the query image based on the proposal parameters. Each cropped image and the reference image are then encoded into feature descriptors in a high-dimensional space. To this end, we employ a pretrained DINOv2~\cite{oquab2023dinov2} as the feature encoder and perform spatial average pooling over the output feature map to obtain the descriptor. The cosine similarity between the reference and the proposal is then computed as 
\begin{align}
\label{eq:detection_sim}
\hat{s}_i = \frac{\mathbf{f}_r\cdot\mathbf{f}_i}{{{||\mathbf{f}_r||}_2{||\mathbf{f}_i||}_2}}, \ \ \mathbf{f}_r, \mathbf{f}_i\in\mathbb{R}^{C_d}
\end{align}
where $\mathbf{f}_r$ and $\mathbf{f}_i$ denote the feature descriptors of the reference and a proposal. We select the proposal with the highest cosine similarity as the final detection result. It is worth noting that one plausible alternative is to utilize the confidence score $s_i$ predicted via the open-vocabulary object detector. However, as will be demonstrated in Section~\ref{sec:experiments}, such an alternative is less effective than our retrieval-based strategy.

\subsection{Hypothesis-Free Relative Rotation Estimation}
Given the bounding box predicted by our open-set object detection method, we crop the object from the query image. In this section, we focus on estimating the relative object rotation between the cropped query image and the reference image.
\begin{figure*}[!t]
    \centering	
    \includegraphics[width=0.9\linewidth]{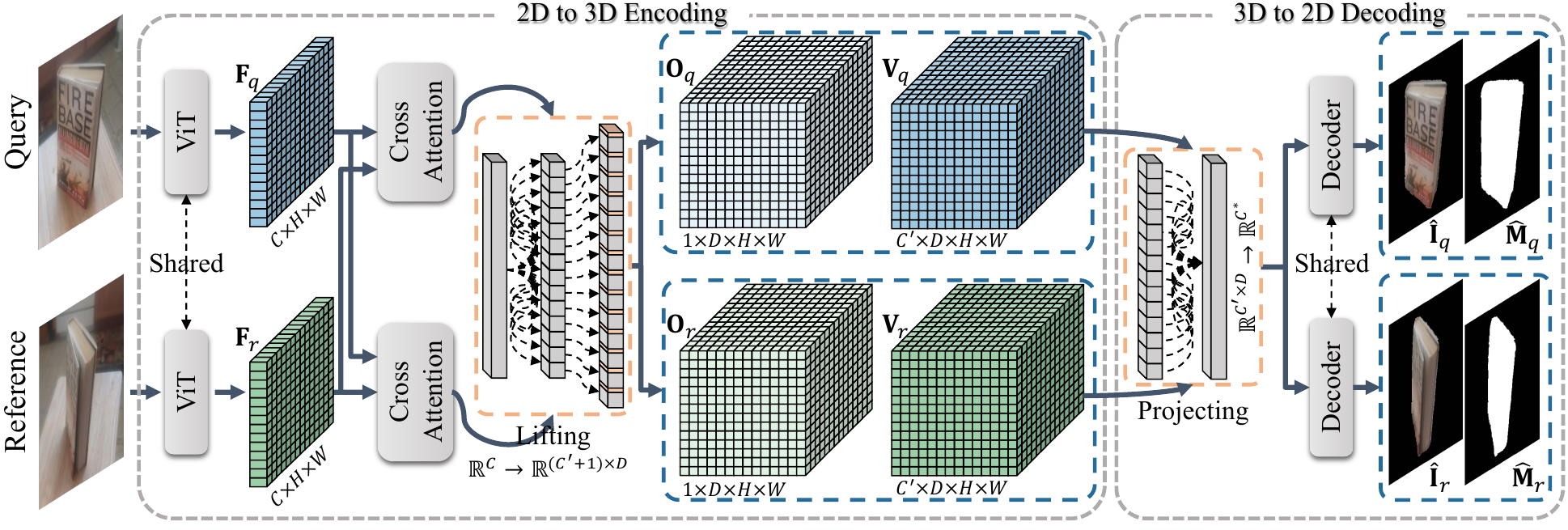}
    \caption{\textbf{Network architecture of our autoencoder.} The encoder takes two RGB images, query and reference, as input and lifts their 2D feature embeddings to 3D voxels by leveraging cross-view 3D information. $\mathbf{O}_q$ and $\mathbf{O}_r$ represent the learned 3D objectness maps account for robust object rotation estimation. The decoder then reconstructs the masked object images from the voxels, allowing the voxels to encode the object patterns.}
    \label{fig:network_ae}
\end{figure*}

\subsubsection{Motivation} 
Drawing inspiration from the success of pixel-pixel correspondences in image matching~\cite{yi2018learning,sarlin2020superglue,zhao2021progressive}, a natural approach to avoiding the use of rotation hypotheses would be to compute the relative object rotation based on 2D correspondences. However, recent studies~\cite{lin2023relpose++,zhao20233d} have observed that such an image-matching strategy is unreliable in the scenario of object pose estimation. We trace this limitation back to the fact that image-matching methods are not fully differentiable w.r.t. the rotation. Specifically, some approaches~\cite{yi2018learning,zhao2019nm,zhang2019learning} encode a notion of consistency among the pixel-pixel correspondences utilizing the essential matrix. However, computing the rotation from the essential matrix leads to multiple solutions~\cite{hartley2003multiple}. Rotation estimation is thus detached from the learning process as a post-processing step. Notably, in the context of object rotation estimation, those pre-generated correspondences tend to be unreliable in the presence of challenges such as large object rotation differences and textureless objects. Therefore, the isolated rotation estimation step in the two-stage design becomes less effective.

To address this issue, we propose to lift the input images to voxelized 3D representations~\cite{wei2020deepsfm} and perform the matching process in 3D latent space. Therefore, the computation of the relative object rotation from the resulting voxel-voxel matches becomes a differentiable operation. This characteristic enables us to directly supervise the rotation estimation module with the actual quantity we aim to predict, i.e., the relative object pose. Below, we elaborate on the steps involved in the presented hypothesis-free mechanism.

\subsubsection{Image Voxelization}
To achieve object rotation estimation from voxel-voxel correspondences with only RGB images as input, we first need to lift each RGB image to a set of 3D voxels. To enable such a voxelization, we introduce an autoencoder network depicted in Fig.~\ref{fig:network_ae}, which includes a 2D-3D encoder and a 3D-2D decoder. Specifically, we employ a pretrained vision transformer~\cite{weinzaepfel2023croco} to convert the query and reference images to 2D feature embeddings denoted as $\mathbf{F}_q$ and $\mathbf{F}_r$, respectively. Considering the difficulty of lifting 2D images to 3D representations, we incorporate a cross-attention module to capture cross-view 3D information. We take the feature embedding $\mathbf{F}_q$ as an example (a symmetric process is carried out for $\mathbf{F}_r$). The cross-attention module~\cite{weinzaepfel2023croco} is defined as
\begin{align}
\label{eq:cross}
& \Tilde{\mathbf{F}}_q^{l}=\text{MHSA}(\text{LN}(\mathbf{F}_q^{l-1})) + \mathbf{F}_q^{l-1}, \\
& \hat{\mathbf{F}}_q^{l}=\text{MHCA}(\text{LN}(\Tilde{\mathbf{F}}_q^{l}), \text{LN}(\mathbf{F}_r^{l-1})) + \Tilde{\mathbf{F}}_q^{l}, \\
& \mathbf{F}_q^{l}=\text{FFN}(\text{LN}(\hat{\mathbf{F}}_q^{l})) + \hat{\mathbf{F}}_q^{l},
\end{align}
where MHSA stands for a multi-head self-attention layer, MHCA represents a multi-head cross-attention layer that takes $\Tilde{\mathbf{F}}_q^{l}$ as \emph{query} and $\mathbf{F}_r^{l-1}$ as \emph{key} and \emph{value}, LN denotes layer normalization~\cite{ba2016layer}, and FFN is a feed-forward network that includes MLPs. The resulting $\hat{\mathbf{F}}_q^{l}$ then serves as the input to the next cross-attention module. Consequently, the output of the last cross-attention module contains object features depicted from two different viewpoints, thus incorporating 3D information. 

Benefiting from such a 3D-aware encoding process, we voxelize the image feature embeddings via a simple reshaping process. 
Note that, to facilitate the robust rotation estimation that will be introduced in Section~\ref{sec:wcv}, we predict an objectness score for each voxel, which reflects the significance of the voxel to the relative object rotation estimation. Therefore, the actual reshaping process is conducted as $\mathbb{R}^{C\times H \times W}\rightarrow \mathbb{R}^{(C^{'}+1)\times D \times H \times W}$, where $C=(C^{'}+1)\times D$. As shown in Fig.~\ref{fig:network_ae}, we denote the resulting 3D objectness maps and 3D volumes as $\mathbf{O}_q, \mathbf{O}_r\in \mathbb{R}^{1 \times D \times H \times W}$ and $\mathbf{V}_q, \mathbf{V}_r \in \mathbb{R}^{C^{'} \times D \times H \times W}$, respectively. Since our approach does not rely on object segmentation, the learned voxel representations may be affected by the background of the query and reference images. To alleviate this issue, we introduce an object-aware decoding process over $\mathbf{V}_q$ and $\mathbf{V}_r$. Concretely, $\mathbf{V}_q$ and $\mathbf{V}_r$ are projected to 2D space by aggregating the voxels along the depth direction as $\mathbb{R}^{C^{'}\times D \times H \times W} \rightarrow  \mathbb{R}^{C^{*}\times H \times W}$, where $C^{*}=C^{'}\times D$. The resulting 2D feature embeddings are then fed into a decoder that contains several self-attention modules~\cite{dosovitskiy2020image} from which the object images $\hat{\mathbf{I}}_q$ and $\hat{\mathbf{I}}_r$ without background are produced. The object masks $\hat{\mathbf{M}}_q$ and $\hat{\mathbf{M}}_r$ are additionally predicted to provide auxiliary information that benefits the following robust object rotation estimation. 

We supervise the training of the autoencoder with an image-level loss function defined as
\begin{align}
\label{eq:img}
& L_{ae} = L_{img} + L_{mask}, \\
& L_{img} = L_{mse}(\hat{\mathbf{I}}_q, \hat{\mathbf{I}}_q^{gt}) + L_{mse}(\hat{\mathbf{I}}_r, \hat{\mathbf{I}}_r^{gt}), \\
& L_{mask} = L_{bce}(\hat{\mathbf{M}}_q, \hat{\mathbf{M}}_q^{gt}) + L_{bce}(\hat{\mathbf{M}}_r, \hat{\mathbf{M}}_r^{gt}), 
\end{align}
where $L_{mse}$ is the mean squared error loss, $L_{bce}$ indicates the binary cross entropy loss, $(\hat{\mathbf{I}}_q^{gt}, \hat{\mathbf{I}}_r^{gt})$ denote the ground-truth foreground images, and $(\hat{\mathbf{M}}_q^{gt}, \hat{\mathbf{M}}_r^{gt})$ represent the ground-truth object masks.

\subsubsection{Object Rotation from Deep Voxel Matching}
\label{sec:dvm}
\begin{figure*}[!t]
    \centering	
    \includegraphics[width=0.9\linewidth]{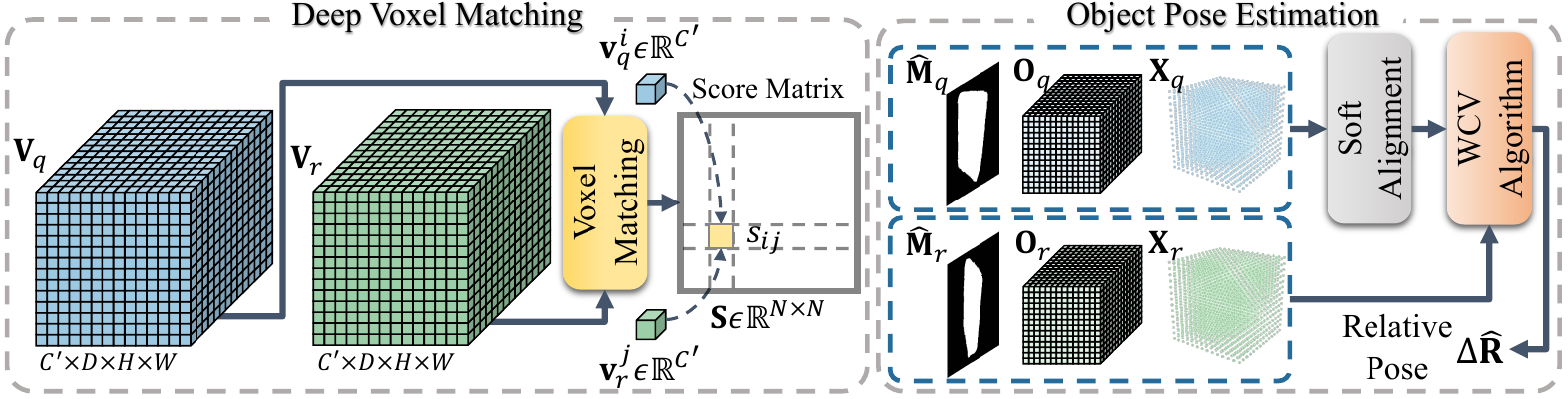}
    \caption{\textbf{Computing relative object rotation from 3D voxels.} The feature similarities of $\mathbf{V}_q$ and $\mathbf{V}_r$ are computed, which results in a score matrix $\mathbf{S}$. A soft assignment is performed based on $\mathbf{S}$ over the query object mask $\hat{\mathbf{M}}_q$, the 3D objectness map $\mathbf{O}_q$, and the 3D coordinates $\mathbf{X}_q$. The aligned query and reference voxels are then fed into a Weighted Closest Voxel (WCV) algorithm that estimates the relative object rotation in a robust and end-to-end manner.}
    \label{fig:wcv}
\end{figure*}

According to multi-view geometry~\cite{besl1992method,hartley2003multiple,wang2019deep}, relative object rotation can be computed by solving a least-squares problem expressed in terms of voxel-voxel correspondences. Specifically, the least-squares problem is formulated as 
\begin{align}
\label{eq:target}
E(\Delta{\mathbf{R}}) = \frac{1}{N}\sum_{i=1}^{N}{||\Delta{\mathbf{R}}\mathbf{x}_{r}^{i}-\mathbf{x}_{q}^{i}||}_2,
\end{align}
where $\mathbf{x}_{r}^{i}\in\mathbf{X}_{r}$ and $\mathbf{x}_{q}^{i}\in\mathbf{X}_{r}$ stand for the 3D coordinates of the $i$-th reference and query voxels, respectively. The coordinates are normalized to be zero-centered and unit-scale. The optimal $\Delta{\hat{\mathbf{R}}}$ is then determined as
\begin{align}
\label{eq:tr}
\Delta{\hat{\mathbf{R}}} = \underset{{\Delta{\mathbf{R}_{i}}\in SO(3)}}{\arg\min}-2\sum_{i=1}^{N}{\mathbf{x}_q^{i}}^{T}\Delta{\mathbf{R}_{i}}\mathbf{x}_r^{i},
\end{align}
As suggested in~\cite{besl1992method}, this problem can be solved by performing a singular value decomposition (SVD) of a covariance matrix as
\begin{align}
\label{eq:svd}
& \mathbf{H}=\sum_{i=1}^{N}\mathbf{x}_r^{i}{\mathbf{x}_q^{i}}^{T}, \\
& \mathbf{H}=\mathbf{U}\mathbf{\Sigma}\mathbf{V}^{T},
\end{align}
where $\mathbf{H}$ indicates the covariance matrix. The closed-form solution to the least-squares problem is given by $\Delta{\hat{\mathbf{R}}}=\mathbf{V}\mathbf{U}^{T}$. Consequently, the key aspect of this problem is to align the 3D voxel coordinates $\mathbf{X}_{q}$ with $\mathbf{X}_{r}$.

Inspired by the studies~\cite{wang2019deep, hu2020single, lin2022category} showing that object pose estimation benefits from end-to-end training, we carry out the alignment in a differentiable fashion. As illustrated in Fig.~\ref{fig:wcv}, the alignment is conducted based on a deep voxel matching module. Specifically, we compute a score matrix $\mathbf{S}$ whose entry $s_{ij}$ indicates the cosine similarity between two voxels as 
\begin{align}
\label{eq:cosine}
s_{ij} = \frac{\mathbf{v}_q^{i}\cdot\mathbf{v}_r^{j}}{{||\mathbf{v}_q^{i}||}_2{||\mathbf{v}_r^{j}||}_2},
\end{align}
where $\mathbf{v}_q^{i}\in\mathbb{R}^{C^{'}}$ and $\mathbf{v}_r^{j}\in\mathbb{R}^{C^{'}}$ denote the $i$-th voxel in $\mathbf{V}_q$ and the $j$-th voxel in $\mathbf{V}_r$, respectively.
The alignment is then achieved as
\begin{align}
\label{eq:align}
\mathbf{X}_q^{'}=p(\mathbf{S}/\tau)\mathbf{X}_q,
\end{align}
where $p(\cdot)$ represents the softmax process and $\tau$ is a predefined temperature.

\subsubsection{Weighted Closest Voxel Algorithm}
\label{sec:wcv}

\begin{figure}[!t]
    \centering	
    \includegraphics[width=1.0\linewidth]{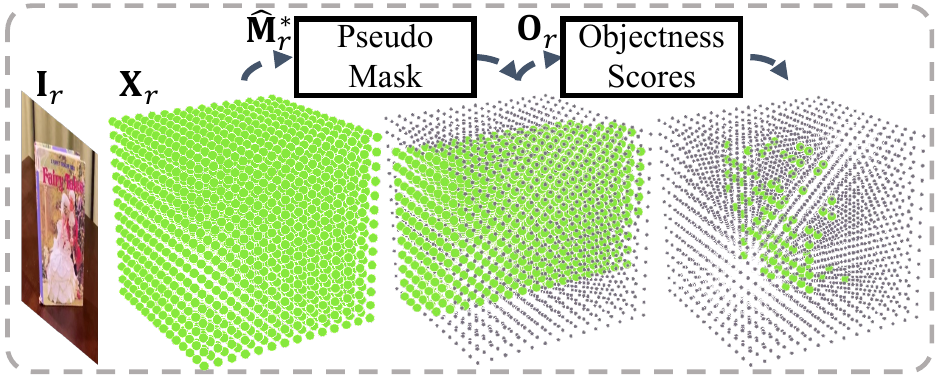}
    \caption{\textbf{Illustration of the voxel weights.} Each colored dot indicates the voxel position in 3D space. The green dots in the middle and right cubes represent the voxels with larger weights. The voxel weights in the middle cube are computed based on the replicated object mask. The weights in the right cube are updated by integrating the 3D objectness map with the object mask.}
    \label{fig:weight}
\end{figure}

Note that our task differs from standard point cloud registration~\cite{segal2009generalized,wang2019deep,aoki2019pointnetlk}, which typically operates on 3D point clouds sampled from 3D object meshes~\cite{chang2015shapenet} or captured using specific sensors~\cite{dai2017scannet}. Here, by contrast, we work with 3D volumes lifted from 2D images, and some voxels could thus be outliers since the corresponding 2D image patches may depict nuisances such as the background. The presence of these outliers may impact the accuracy of the relative object rotation estimated from the voxel matches. To address this challenge, we introduce a weighted closest voxel algorithm that enables robust relative object rotation estimation.

Concretely, the objective is to mitigate the effect of unreliable voxel matches. We thus incorporate a weight vector into the rotation estimation process, modifying Eq.~\ref{eq:svd} as
\begin{align}
\label{eq:wsvd}
\mathbf{H}=\sum_{i=1}^{N}\omega^{i}\mathbf{x}_r^{i}{\mathbf{x}_q^{i}}^{T},
\end{align}
where $\omega^{i}\in (0, 1)$ denotes the weight of the $i$-th voxel pair. This makes the subsequent relative object rotation estimation aware of the reliability of each voxel pair. We determine the weight vector by utilizing both the object mask and voxel objectness information produced by the encoder network. Specifically, we first replicate $\mathbf{\hat{M}}_q$ and $\mathbf{\hat{M}}_r$ $D$ times along the depth dimension, which creates \emph{pseudo} 3D masks, $\mathbf{\hat{M}}_q^{*}, \mathbf{\hat{M}}_r^{*}\in \mathbb{R}^{1 \times D \times H \times W}$. These pseudo 3D masks contribute to alleviating the influence of voxels that depict the background. The weight of each voxel pair is then determined as 
\begin{align}
\label{eq:img_weight}
\mathbf{W}_m = h(\frac{p(\mathbf{S}/\tau)\mathbf{\hat{M}}_q^{*}+\mathbf{\hat{M}}_r^{*}}{2\lambda}),
\end{align}
where $h(\cdot)$ indicates the sigmoid function, and $\lambda$ is a manually defined temperature. Additionally, to mitigate the redundancies naturally introduced by the replication process over $\mathbf{\hat{M}}_q$ and $\mathbf{\hat{M}}_r$, we integrate the resulting pseudo masks with the 3D objectness maps. The final weight vector of all pairwise voxels is determined as $\mathbf{W}=\mathbf{W}_{o}\odot\mathbf{W}_{m}$, where $\odot$ indicates the Hadamard product, and $\mathbf{W}_{o}$ is obtained by carrying out Eq.~\ref{eq:img_weight} over $\mathbf{O}_q$ and $\mathbf{O}_r$. 

Fig.~\ref{fig:weight} provides an example of the estimated voxel weights. The dots denote the 3D voxel positions and the voxels assigned with larger weights are colored in green within the middle and right cubes. In the right cube, the green dots roughly depict a 3D surface that corresponds to the object visible in the 2D image. Note that our rotation estimation network is trained without relying on ground-truth 3D object models. This observation thus demonstrates that the voxels that are crucial in determining the relative object rotation are aware of the 3D object shape information.
The complete rotation estimation module is trained end-to-end with a loss function defined as $L=L_{ae} + L_{pose}$ with 
\begin{align}
\label{eq:pose}
L_{pose}=||q(\Delta{\hat{\mathbf{R}}}) - q(\Delta\mathbf{R}^{gt})||,
\end{align}
where $\Delta\mathbf{R}^{gt}$ is the ground-truth relative object rotation, and $q(\cdot)$ is a function that converts a rotation matrix to a 6D continuous representation~\cite{zhou2019continuity}.

%% file: sec/4_experiments.tex
\section{Experiments}
\label{sec:experiments}
\begin{table*}[!t]
\small
    \caption{\textbf{Relative object rotation estimation on CO3D~\cite{reizenstein2021common}.} We report the angular errors of the estimated relative object rotations. All testing object categories were unseen during training. The best results are shown in bold fonts.}
    \begin{center}
        \begin{tabular}{lccccccccccc}
        \Xhline{2\arrayrulewidth}
        Method & Ball & Book & Couch & Frisb. & Hotd. & Kite & Remot. & Sandw. & Skate. & Suitc. & \textbf{Mean} \\
        \hline 
        SG~\cite{sarlin2020superglue} & 83.55 & 71.02 & 45.14 & 68.67 & 88.74 & 56.46 & 78.58 & 73.64 & 72.14 & 76.74 & 71.47 \\
        LoFTR~\cite{sun2021loftr} & 82.51 & 77.33 & 60.57 & 78.39 & 85.05 & 70.03 & 89.74 & 77.77 & 74.33 & 90.73 & 78.64 \\
        ZSP~\cite{goodwin2022zero} & 88.09 & 90.09 & 64.07 & 79.08 & 99.62 & 72.71 & 98.61 & 89.09 & 89.41 & 95.03 & 86.66 \\
        Regress~\cite{lin2023relpose++} & 47.56 & 52.91 & 39.12 & 50.16 & 51.28 & 52.33 & 43.85 & 52.89 & 51.59 & 29.11 & 47.08 \\
        RelPose~\cite{zhang2022relpose} & 56.96 & 55.89 & 40.71 & 54.11 & 64.20 & 69.43 & 42.89 & 59.05 & 42.32 & 32.50 & 51.80 \\
        RelPose++~\cite{lin2023relpose++} & 36.42 & 35.64 & 20.00 & 36.27 & 33.62 & 33.63 & 34.83 & 36.93 & 40.60 & 20.32 & 32.82 \\
        PoseDiffusion~\cite{wang2023posediffusion} & 41.38 & 35.05 & 42.41 & 39.64 & 87.16 & 51.35 & 25.09 & 61.64 & 38.46 & 23.66 & 44.58 \\
        3DAHV~\cite{zhao20233d} & 34.83 & 31.21 & 22.12 & 31.30 & 35.39 & 34.96 & 24.73 & 26.97 & 26.81 & \textbf{16.13} & 28.44 \\
        \textbf{DVMNet++} & \textbf{28.31} & \textbf{21.98} & \textbf{19.01} & \textbf{23.23} & \textbf{21.45} & \textbf{17.50} & \textbf{11.39} & \textbf{19.63} & \textbf{20.14} & 16.85 & \textbf{19.95} \\
        \Xhline{2\arrayrulewidth}
        \end{tabular}
    \end{center}
    \label{tab:co3d}
\end{table*}

\subsection{Implementation Details}
In the presented autoencoder, we use 3 cross-attention modules in the 2D-3D encoder and 3 self-attention modules in the 3D-2D decoder. In the relative object rotation estimation module, we normalize the 3D coordinates of the voxels to an interval of $[-1, 1]$ with a mean of $\mathbf{0}$. We set the temperatures $\tau$ and $\lambda$ in Eq.~\ref{eq:align} and Eq.~\ref{eq:img_weight} to $0.1$ and $1.0$, respectively. We train our network on an A100 GPU, employing the AdamW~\cite{loshchilov2017decoupled} optimizer with a
batch size of 64 and a learning rate of $10^{-5}$. We crop the object from the query image using the ground-truth object bounding box during the training stage, following the implementation in~\cite{zhang2022relpose,lin2023relpose++,zhao20233d}. We replace the ground truth with the bounding box predicted by the proposed open-set object detector at test time.

\subsection{Relative Object Rotation Estimation on CO3D}
We first evaluate our method on the CO3D dataset~\cite{reizenstein2021common}, which has been commonly utilized in the literature~\cite{zhang2022relpose,lin2023relpose++,zhao20233d}. This dataset contains 18,619 video sequences that depict 51 object categories. To evaluate the generalization ability of the network to unseen objects, we follow the setting in~\cite{zhang2022relpose}, training the network on 41 object categories and testing it on the other 10 categories. The performance is measured by the mean angular error $err_R\in[0^{\circ}, 180^{\circ}]$ of the estimated relative object rotation, which is defined as 
\begin{align}
\label{eq:rota_error}
err_R = \text{arccos}\left(\frac{\text{tr}(\Delta{\hat{\mathbf{R}}^{T}}\Delta{\mathbf{R}_{gt}})-1}{2}\right).
\end{align}
We compare our approach with state-of-the-art techniques including image-matching methods, SuperGlue (SG)~\cite{sarlin2020superglue}, LoFTR~\cite{sun2021loftr}, and ZSP~\cite{goodwin2022zero}, hypothesis-based methods, RelPose~\cite{zhang2022relpose}, RelPose++~\cite{lin2023relpose++}, and 3DAHV~\cite{zhao20233d}, a diffusion-based method, PoseDiffusion~\cite{wang2023posediffusion}, and a direct regression method implemented in~\cite{lin2023relpose++}. Note that since we focus on the evaluation of relative rotation estimation, we use the ground-truth object bounding box to locate the object in the query image and only utilize a single reference image. We maintain this setting across all evaluated methods to ensure a fair comparison.

\begin{figure}[!t]
    \centering	
    \includegraphics[width=0.9\linewidth]{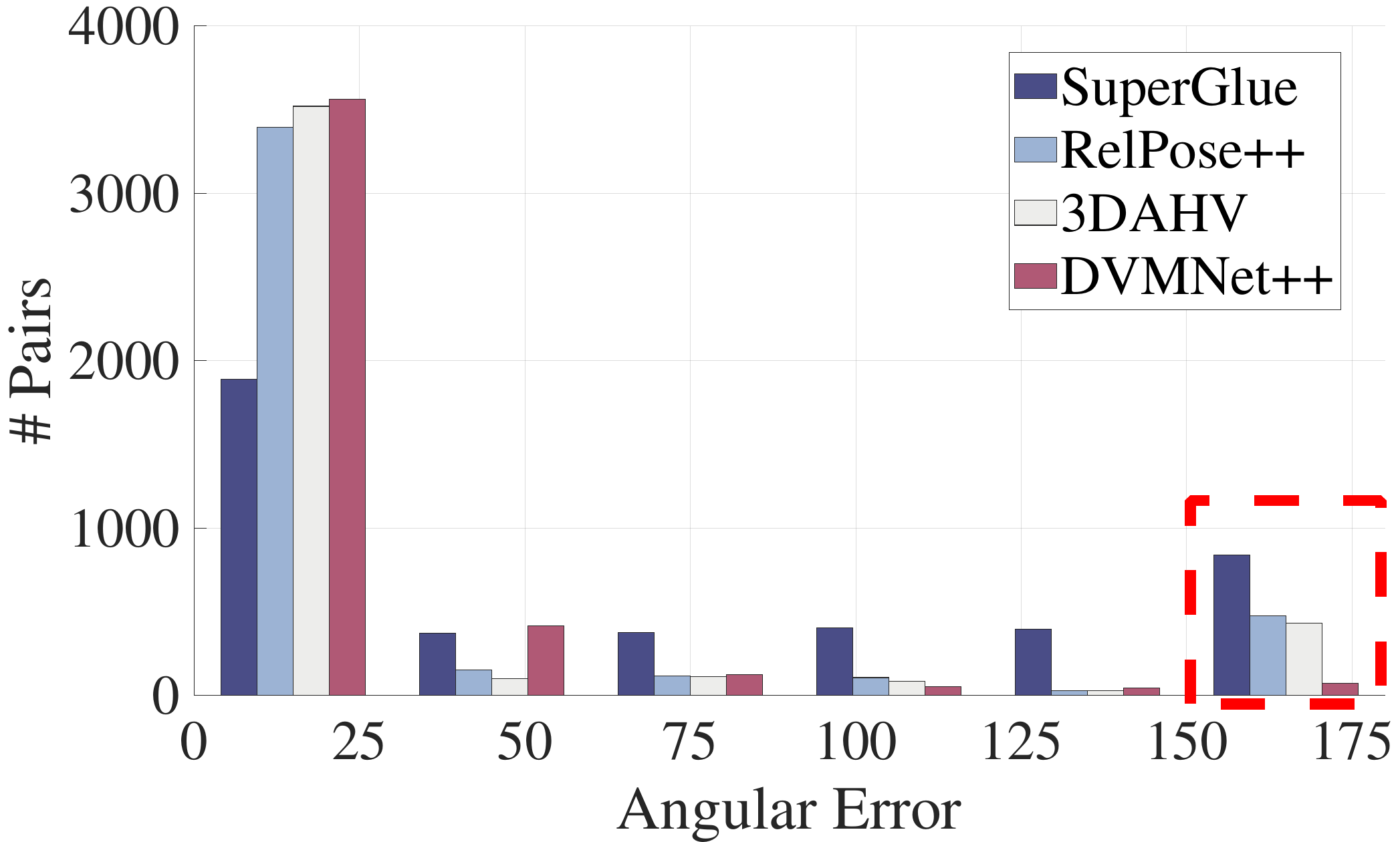}
    \caption{\textbf{Histogram depicting the distribution of rotation errors.} The image pairs in the testing set are divided into distinct groups based on the angular errors obtained by a specific rotation estimation approach. Each bar in the histogram represents the count of image pairs within a particular group. Our DVMNet++ yields much fewer unnaturally large errors than image-matching and hypothesis-based methods.}
    \label{fig:hist}
\end{figure}

As reported in Table~\ref{tab:co3d}, DVMNet++ delivers superior relative rotation estimation performance for unseen objects, outperforming both the image-matching and hypothesis-based competitors by at least $8.49^{\circ}$ in terms of mean angular error. To shed more light on the robustness of the evaluated approaches, we categorize the testing image pairs into different groups according to the corresponding angular errors observed when applying a particular relative object rotation estimation method. We count the number of image pairs in each group and show the results in Fig.~\ref{fig:hist}. Our method results in a higher number of image pairs with smaller angular errors. More importantly, as highlighted by the red dashed box in Fig.~\ref{fig:hist}, both image-matching and hypothesis-based methods exhibit large angular errors for some image pairs. By contrast, our DVMNet++ yields fewer failure instances, thus demonstrating better robustness.

\begin{table}[!t]
    \caption{\textbf{Time consumption.} We evaluate the speed on an A100 GPU. The average time consumption per image pair is reported. For hypothesis-based approaches, the rotation hypotheses are processed in parallel.}
    \begin{center}
        \small
        \begin{tabular}{cccc}
        \Xhline{2\arrayrulewidth}
        RelPose++ & PoseDiffusion & 3DAHV & \textbf{DVMNet++} \\
        \hline 
        29ms & 5584ms & 35ms & 23ms \\
        \Xhline{2\arrayrulewidth}
        \end{tabular}
    \end{center}   
    \label{tab:speed}
\end{table}

\begin{figure}[!t]
    \centering	
    \includegraphics[width=0.9\linewidth]{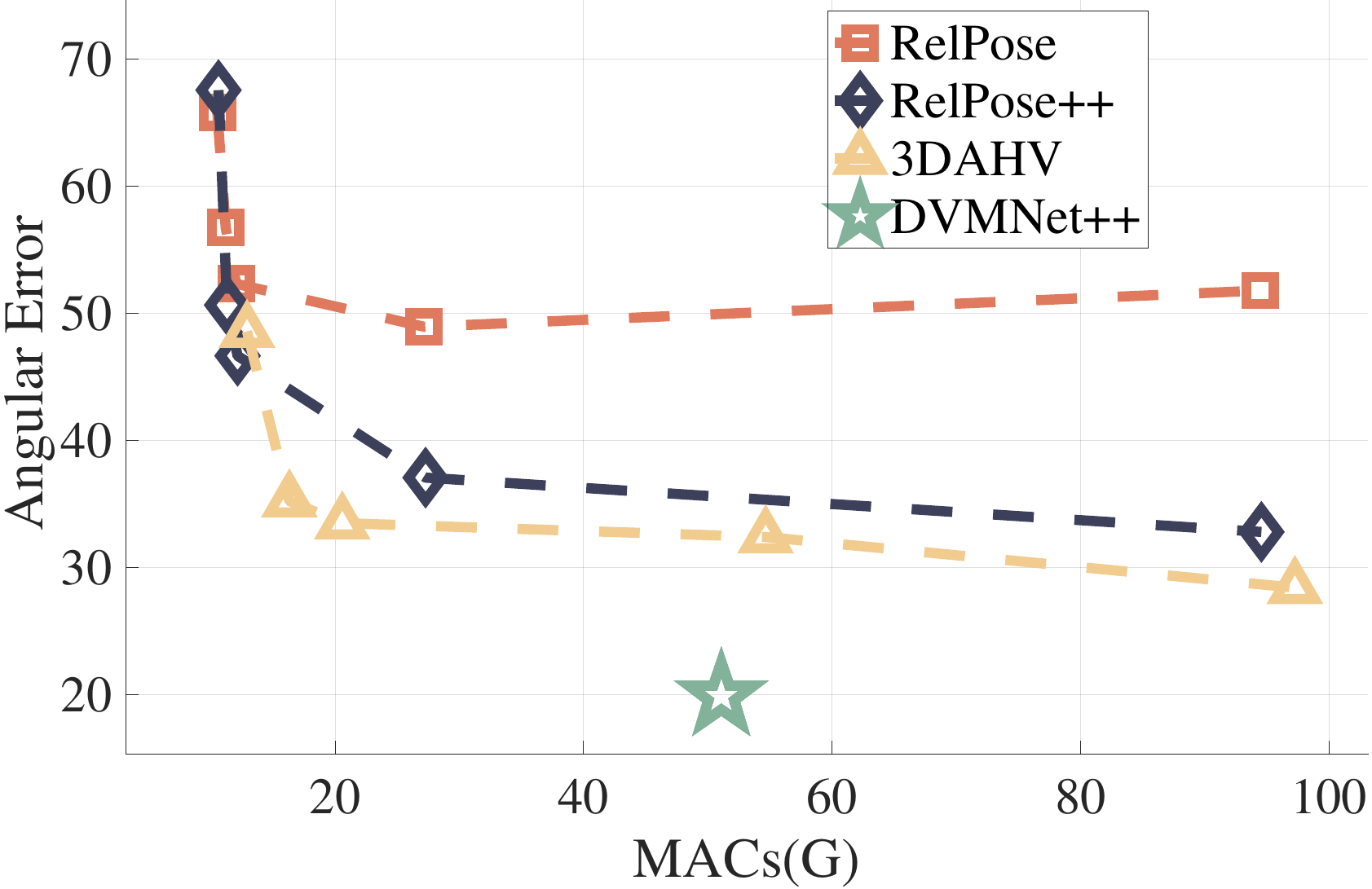}
    \caption{\textbf{Comparison with hypothesis-based methods.} We measure the computational cost as multiply-accumulate operations (MACs). The results for hypothesis-based methods are shown with varying numbers of rotation samples, ranging from 1,000 to 500,000. For 3DAHV, we set the maximum number to be 100,000 due to the computational resource constraints.}
    \label{fig:cost}
\end{figure}

Furthermore, as argued in Section~\ref{sec:method}, our hypothesis-free strategy is more efficient than the hypothesis-based techniques in relative object rotation estimation. We thus assess their computational cost, utilizing the multiply-accumulate operations (MACs). For hypothesis-based methods, all sampled hypotheses are processed in parallel. The results shown in Fig.~\ref{fig:intro} indicate the benefits of our hypothesis-free DVMNet++, which requires considerably fewer MACs than the hypothesis-based competitors. To further substantiate this advantage, we provide detailed results in Fig.~\ref{fig:cost}, where the hypothesis-based methods are evaluated with the number of rotation samples varying from 1,000 to 500,000. Note that for 3DAHV, the maximum number is 100,000 because of our computational resource constraints. As shown in Fig.~\ref{fig:cost}, one can enhance the efficiency of the hypothesis-based methods by reducing the number of samples. However, this efficiency gain comes at the cost of sacrificing rotation estimation accuracy. By contrast, our method achieves a good trade-off between efficiency and rotation estimation accuracy. We also evaluate the time consumption on an A100 GPU and the results are reported in Table~\ref{tab:speed}. On average, DVMNet++ processes a pair of images in~\emph{23ms}. Despite benefiting from parallel estimation, the hypothesis-based methods RelPose++ and 3DAHV are still slower than our method.

\begin{table*}[!t]
    \small
    \caption{\textbf{Relative object rotation estimation on the GROP benchmark~\cite{zhao20233d}.} The methods are evaluated in terms of angular error on the LINEMOD~\cite{hinterstoisser2012model} and Objaverse~\cite{deitke2023objaverse} datasets. The testing data comprises 5 objects from LINEMOD and 128 objects from Objaverse. All images containing these objects are omitted from the training set.}
    \begin{center}
        \begin{tabularx}{\textwidth}{lcccccccc} 
        \Xhline{2\arrayrulewidth}
        \rowcolor{Gray}
        LINEMOD & SG~\cite{sarlin2020superglue} & LoFTR~\cite{sun2021loftr} & ZSP~\cite{goodwin2022zero} & Regress~\cite{lin2023relpose++} & RelPose~\cite{zhang2022relpose} & RelPose++~\cite{lin2023relpose++} & 3DAHV~\cite{zhao20233d} & \textbf{DVMNet++} \\
        \hline 
        Cat & 67.28 & 88.06 & 79.61 & 54.21 & 53.72 & 47.77 & 50.99 & \textbf{31.70} \\
        Ben. & 58.52 & 70.80 & 74.07 & 52.03 & 62.32 & 44.67 & 38.16 & \textbf{34.00} \\
        Cam. & 58.11 & 87.13 & 79.65 & 51.04 & 59.91 & 44.31 & 41.92 & \textbf{33.18} \\
        Dri. & 65.16 & 78.85 & 76.35 & 52.83 & 57.61 & 47.95 & \textbf{32.65} & 46.29 \\
        Duck & 74.90 & 97.63 & 83.43 & 55.44 & 55.15 & 48.65 & 44.03 & \textbf{38.91} \\
        \textbf{Mean} & 64.79 & 84.49 & 78.62 & 53.11 & 57.75 & 46.67 & 41.55 & \textbf{36.82} \\
        \Xhline{2\arrayrulewidth}
        \rowcolor{Gray}
        Objaverse & SG~\cite{sarlin2020superglue} & LoFTR~\cite{sun2021loftr} & ZSP~\cite{goodwin2022zero} & Regress~\cite{lin2023relpose++} & RelPose~\cite{zhang2022relpose} & RelPose++~\cite{lin2023relpose++} & 3DAHV~\cite{zhao20233d} & \textbf{DVMNet++} \\
        \hline 
        \textbf{Mean} & 102.40 & 134.05 & 107.20 & 55.90 & 80.39 & 33.49 & 28.11 & \textbf{20.19} \\
        \Xhline{2\arrayrulewidth}
        \end{tabularx}
    \end{center}
    \label{tab:grop}
\end{table*}

\subsection{Relative Object Rotation Estimation on GROP}
Recently, a new benchmark called GROP for relative rotation estimation of unseen objects was introduced in~\cite{zhao20233d}. This benchmark comprises two datasets, i.e., Objaverse~\cite{deitke2023objaverse} and LINEMOD~\cite{hinterstoisser2012model}. Both synthetic and real images with diverse object poses are considered. We perform experiments on these two datasets, following the same setup as described in~\cite{zhao20233d}. More concretely, the synthetic images are generated by rendering the object models of the Objaverse dataset from different viewpoints~\cite{liu2023zero}. Several sequences of calibrated real images that depict 13 texture-less household objects are provided from the LINEMOD dataset. The testing set encompasses 128 objects from Objaverse and 5 objects from LINEMOD. The images containing these objects are excluded from the training data, ensuring that all testing objects are previously unseen. All evaluated approaches are trained and tested on the same predefined image pairs, leading to a fair comparison. As  in~\cite{zhao20233d}, we crop the object from the query image employing the ground-truth object bounding boxes.

\begin{figure}[!t]
    \centering	
    \includegraphics[width=1.0\linewidth]{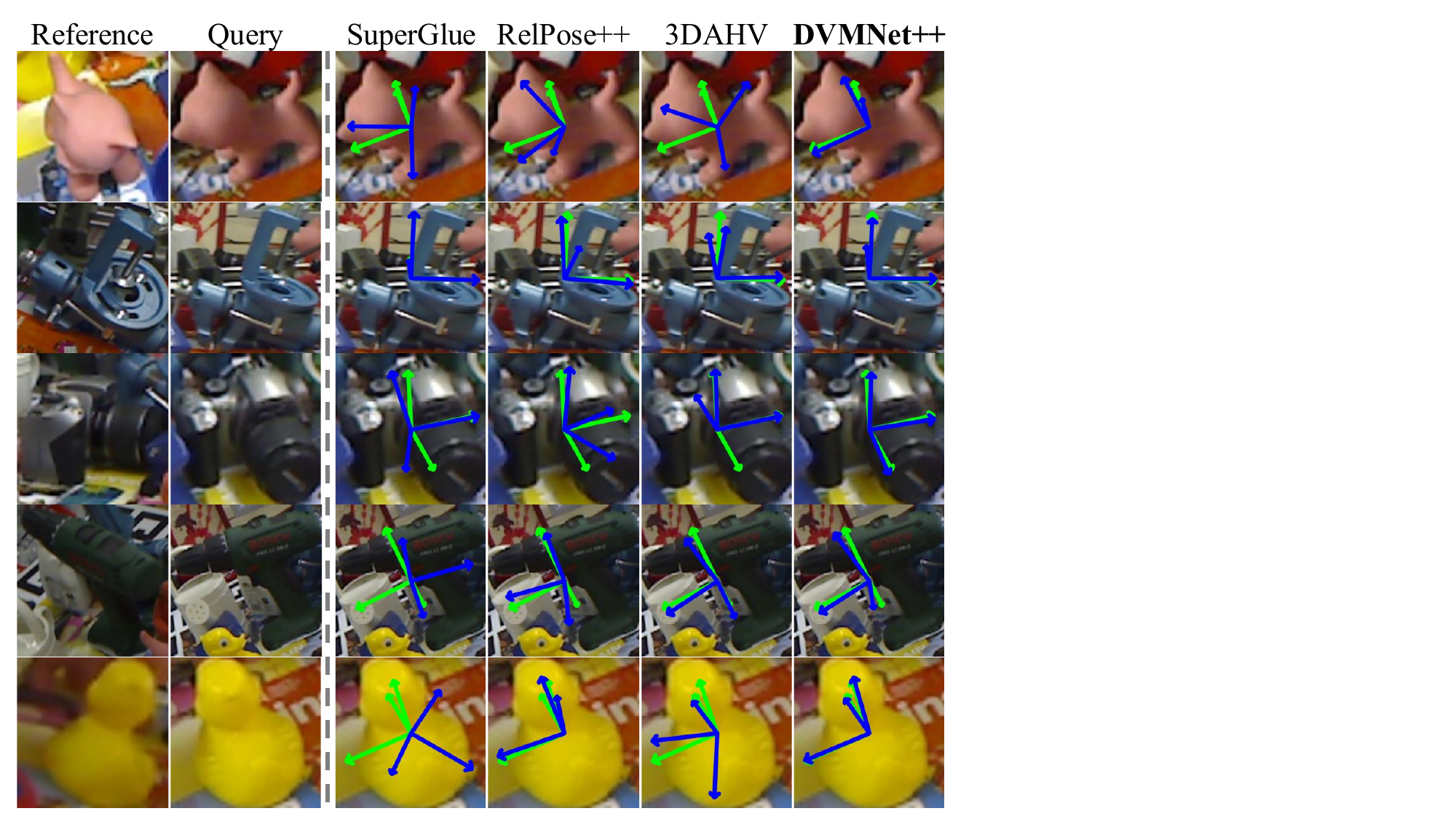}
    \caption{\textbf{Qualitative rotation estimation results on LINEMOD~\cite{hinterstoisser2012model}.} We visualize the object rotation in the query image based on the relative object rotation. The green and blue arrows represent the ground-truth object rotations and the estimated ones, respectively.} 
    \label{fig:viz}
\end{figure}

Table~\ref{tab:grop} provides the angular errors of the estimated relative object rotations on the LINEMOD and Objaverse datasets. In the synthetic scenarios of Objaverse, DVMNet++ outperforms the previous methods by at least $7.92^{\circ}$ in terms of mean angular error. In the real scenarios of LINEMOD, DVMNet++ achieves the smallest angular error for most of the testing objects and reduces the mean angular error by at least $4.73^{\circ}$ compared to the other approaches. Moreover, we visualize the object rotation depicted in the query image and show qualitative results in Fig.~\ref{fig:viz}. The query object rotation is determined as $\mathbf{R}_q=\Delta\mathbf{R}\mathbf{R}_r$. The ground-truth and predicted query object rotations are represented as green and blue arrows, respectively. It is evident from Fig.~\ref{fig:viz} that the rotations estimated with our DVMNet++ are consistently more similar to the ground truth than those obtained with the baselines.

\subsection{Relative Object Pose with Open-Set Detector}
Recall that, in the preceding experiments, we use the ground-truth object bounding box to crop the object from the query image, which aligns with the setting in~\cite{zhang2022relpose,lin2023relpose++,zhao20233d}. To evaluate the effectiveness of the proposed open-set object detector, we conduct experiments on CO3D and LINEMOD, replacing the ground-truth bounding boxes with the predicted ones. We use the object category as the text prompt when performing the object detection on CO3D. Since LINEMOD contains cluttered scenarios, unseen object detection is more challenging. Therefore, we incorporate additional attribute descriptions, such as color and material, based on the reference image. 

\begin{figure*}[!t]
    \centering	
    \includegraphics[width=1.0\linewidth]{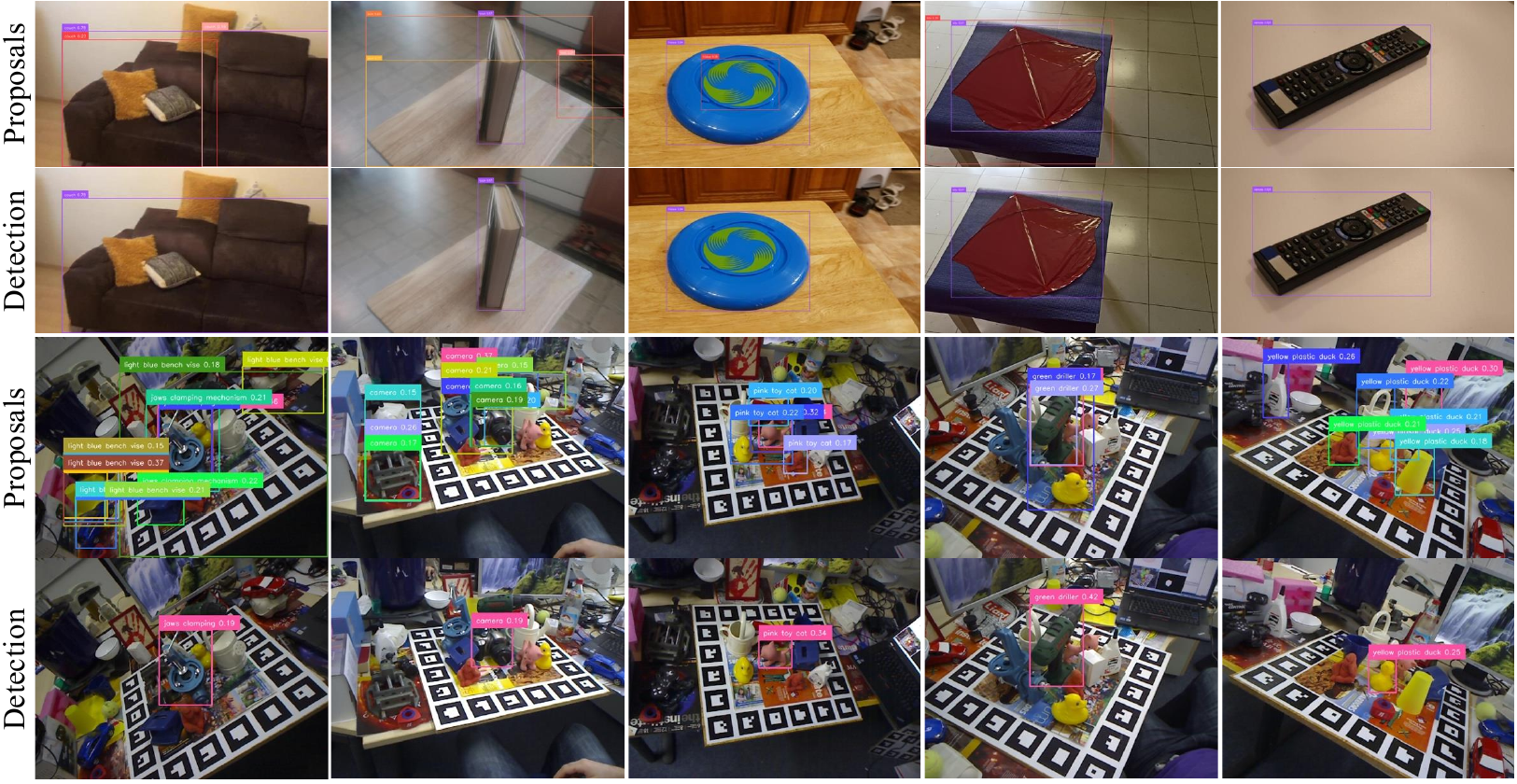}
    \caption{\textbf{Qualitative unseen object detection results on CO3D~\cite{reizenstein2021common} and LINEMOD~\cite{hinterstoisser2012model}.} The object proposals are obtained by employing the open-vocabulary object detector~\cite{liu2023grounding}. We determine the final detection result from the proposals using the introduced image retrieval strategy.} 
    \label{fig:dect_viz}
\end{figure*}

We evaluate our method in terms of translation estimation error (Trans. Error) and rotation estimation error (Rota. Error). Specifically, we obtain the object bounding box parameters using the presented object detector. The relative object translation is computed based on Eq.~\ref{eq:translation}. The relative object rotation is estimated using the cropped query image and the reference image. To eliminate the scale ambiguity, we compute the angular error~\cite{wang2023posediffusion} of the translation, which is defined as
\begin{align}
\label{eq:trans_err}
err_T = \frac{\Delta\hat{\mathbf{T}}\cdot\Delta\mathbf{T}_{gt}}{{||\Delta\hat{\mathbf{T}}||}_2{||\Delta\mathbf{T}_{gt}||}_2},
\end{align}
where $\Delta\hat{\mathbf{T}}$ and $\Delta\mathbf{T}_{gt}$ denote the predicted relative object translation and the ground truth, respectively. The experimental results on CO3D and LINEMOD are reported in Table~\ref{tab:co3d_6d} and Table~\ref{tab:lm_6d}, respectively. Our open-set object detector yields highly accurate relative object translation estimates, with a mean translation error of $6.47^{\circ}$ on CO3D and $1.55^{\circ}$ on LINEMOD. This demonstrates that, on average, the predicted bounding box center is close to the ground-truth object center in the query image. Moreover, the rotation error based on the predicted object bounding boxes is comparable to those obtained with the ground-truth object positions. This observation highlights the effectiveness of our approach in two aspects: First, our deep voxel matching network is robust to noise in object detection; second, our open-set object detector provides reliable object position parameters. We showcase some detection results in Fig.~\ref{fig:dect_viz}. As shown in this figure, the object proposals generated based on the natural language understanding are noisy, particularly in cluttered scenarios. Our detector building upon multiple modalities is capable of identifying the correct bounding box from the candidates. 

We also evaluate the efficiency of the open-set object detector. On average, the detection takes $\emph{92}ms$ on the CO3D dataset. This demonstrates that our method delivers reliable object detection results for relative object pose estimation with acceptable computational overhead.  

\begin{table*}[!t]
    \caption{\textbf{Unseen object detection results on CO3D~\cite{reizenstein2021common}.} We replace the ground-truth object bounding boxes with the ones predicted by our open-set object detector. The translation estimation errors (Tran. Error) and rotation estimation errors (Rota. Error) with the predicted object bounding boxes are reported.}
    \begin{center}
        \normalsize
        \begin{tabular}{lccccccccccc}
        \Xhline{2\arrayrulewidth}
        CO3D & Ball & Book & Couch & Frisb. & Hotd. & Kite & Remot. & Sandw. & Skate. & Suitc. & \textbf{Mean} \\
        \hline 
        Tran. Error & 4.23 & 3.44 & 3.05 & 10.33 & 2.58 & 3.13 & 4.62 & 2.63 & 2.63 & 3.38 & 6.47 \\
        Rota. Error & 27.66 & 22.04 & 19.44 & 22.69 & 19.98 & 16.67 & 11.47 & 18.91 & 20.25 & 17.25 & 19.64 \\      
        \Xhline{2\arrayrulewidth}
        \end{tabular}
    \end{center}
    \label{tab:co3d_6d}
\end{table*}

\begin{table}[!t]
    \caption{\textbf{Unseen object detection results on LINEMOD~\cite{hinterstoisser2012model}.} We maintain the same experimental settings as those used on CO3D. We show the translation and rotation estimation errors, using the proposed object detector in our pipeline.}
    \begin{center}
        \small
        \begin{tabular}{p{1.6cm}C{0.7cm}C{0.7cm}C{0.7cm}C{0.7cm}C{0.7cm}C{0.7cm}}
        \Xhline{2\arrayrulewidth}
        LINEMOD & Cat & Ben. & Cam. & Dri. & Duck & \textbf{Mean} \\
        \hline 
        Tran. Error & 1.04 & 1.93 & 1.44 & 1.64 & 1.67 & 1.55 \\
        Rota. Error & 33.62 & 38.56 & 35.22 & 48.31 & 43.24 & 39.79 \\      
        \Xhline{2\arrayrulewidth}
        \end{tabular}
    \end{center}   
    \label{tab:lm_6d}
\end{table}

\subsection{Ablation Studies}

\subsubsection{Weighted Closest Voxel Algorithm}
\begin{table}[!t]
    \caption{\textbf{Effectiveness of our WCV algorithm.} We report the mean angular errors of relative object rotation estimation on the CO3D~\cite{reizenstein2021common} dataset. The second row indicates the scenario where the WCV algorithm is replaced with a rotation regression module. The third row presents the closest voxel algorithm without weights involved.}
    \begin{center}
        \normalsize
        \begin{tabular}{cccc}
        \Xhline{2\arrayrulewidth}
        WCV & 2D Mask & Voxel Objectness & Angular Error \\
        \hline 
        \xmark & \xmark & \xmark & 31.78 \\
        \cmark & \xmark & \xmark & 21.64 \\
        \cmark & \cmark & \xmark & 20.92 \\
        \cmark & \xmark & \cmark & 20.07 \\
        \cmark & \cmark & \cmark & 19.95 \\
        \Xhline{2\arrayrulewidth}
        \end{tabular}
    \end{center}
    \label{tab:mask}
\end{table}

As a critical component of our deep voxel matching network, the weighted closest voxel (WCV) algorithm plays a pivotal role in achieving hypothesis-free and end-to-end relative object rotation estimation. To substantiate the effectiveness of the WCV algorithm, we develop comprehensive ablation studies on the CO3D dataset. We first replace the WCV algorithm with a rotation regression module. More concretely, we perform global average pooling over $\mathbf{V}_q$ and $\mathbf{V}_r$. The resulting feature embeddings are concatenated and passed through three fully connected layers to predict the 6D continuous representation of relative object rotation. We maintain all the other components in our framework unchanged to ensure a fair comparison. This alternative approach is also able to predict the relative object rotation in a hypothesis-free and end-to-end fashion. However, as shown in Table~\ref{tab:mask}, the mean angular error on the CO3D dataset increases by more than $10^{\circ}$ when the regression module is employed, showcasing the importance of the WCV algorithm in the presented hypothesis-free mechanism. Furthermore, we evaluate three counterparts of the WCV algorithm, i.e., a closest voxel algorithm without weights, a WCV algorithm with only replicated 2D object masks, and a WCV algorithm with only 3D objectness maps. The final weights of the voxel pairs are determined as $\mathbf{W}_m$ and $\mathbf{W}_o$ in the last two counterparts, respectively. The closest voxel algorithm delivers the worst results among these three variants, revealing that the rotation estimation process is affected by the potential outliers. The best performance is achieved by leveraging both 2D object masks and 3D voxel objectness maps, which thus demonstrates the effectiveness of these components in the proposed rotation estimation module.

\begin{table*}[!t]
    \normalsize
    \caption{\textbf{Extension to sparse-view references.} The experiment is conducted on CO3D~\cite{reizenstein2021common} with the number of reference images varying from 1 to 7. We report the angular error between the computed query object rotation and the ground truth.}
    \centering
    \begin{tabular}{lccccccc}
    \Xhline{2\arrayrulewidth}
    \# References & 1     & 2     & 3     & 4     & 5     & 6     & 7     \\
    \hline
    SuperGlue~\cite{sarlin2020superglue} & 71.47 & 73.08 & 68.72 & 65.90 & 64.54 & 63.24 & 61.74 \\
    3DAHV~\cite{zhao20233d} & 28.44 & 29.29 & 28.20 & 27.21 & 26.40 & 24.85 & 24.76 \\
    \textbf{DVMNet++} & \textbf{19.95} & \textbf{18.38} & \textbf{16.79} & \textbf{16.21} & \textbf{15.73} & \textbf{14.99} & \textbf{14.93} \\
    \Xhline{2\arrayrulewidth}
    \end{tabular}
    \label{tab:sparse}
\end{table*}

\subsubsection{Extension to Sparse References}
Note that, by default, we assume that a single reference image is available in our experiments. However, to account for scenarios where multiple reference images may be provided, we conduct an experiment on the CO3D dataset to evaluate the compatibility of our method with such sparse references. Specifically, given an unseen object during testing, we randomly sample $n$ images, with $n$ ranging from 2 to 8. These images are then fed into our network, with one image designated as the query and the remaining ones as references. As shown in the preceding experiments, relative object rotation estimation is more challenging than translation estimation. Consequently, we focus on evaluating rotation estimation in this experiment. The object rotation in the query image is simply derived from the resulting $n-1$ relative object rotations as
\begin{align}
\label{eq:sparse}
\mathbf{R}_q=h^{-1}(\frac{1}{n-1}\sum_{i=1}^{n-1}h(\Delta{\hat{\mathbf{R}}_i} \mathbf{R}_r^{i})),
\end{align}
where $\Delta{\hat{\mathbf{R}}_i}$ and $\mathbf{R}_r^{i}$ denote the $i$-th relative object rotation and reference rotation, respectively, $h(\cdot)$ represents a function that converts a rotation matrix to the 6D continuous representation~\cite{zhou2019continuity}, and $h^{-1}(\cdot)$ indicates the inverse conversion. We also evaluate the representative image-matching (SuperGlue) and hypothesis-based (3DAHV) approaches in the sparse-view scenario. We ensure a fair comparison by utilizing the same strategy of query object rotation estimation for these methods. We report the resulting rotation errors in Table~\ref{tab:sparse}. It is evident that (i) the rotation error of our method decreases as more reference images are involved, and (ii) our method consistently yields the smallest rotation error. These observations demonstrate the promising compatibility of our approach with sparse reference images.

\begin{table}[!t]
    \normalsize
    \caption{\textbf{Effectiveness of the object detector.} The relative object translation is approximated using the parameters of the bounding box. GT indicates that the ground-truth bounding box is used. w/o RGB means the detection result is selected from the proposals using the confidence scores. We report the mean translation errors on LINEMOD~\cite{hinterstoisser2012model}.}
    \centering
    \begin{tabular}{lccc}
    \Xhline{2\arrayrulewidth}
    Method & GT & w/o RGB & \textbf{Ours} \\
    \hline
    Tran. Error & 0.75 & 5.27 & 1.55 \\
    \Xhline{2\arrayrulewidth}
    \end{tabular}
    \label{tab:ana_dete}
\end{table}

\subsubsection{Open-Set Object Detection}
To shed more light on the effectiveness of our open-set object detection module, we conduct experiments on LINEMOD, comparing the method with several alternatives. We first evaluate the method (GT) using the ground-truth object bounding box. Recall that we approximate the 2D projection of the 3D object center in the object coordinate system as the center of the object bounding box. These positions may differ, even when using the ground-truth bounding box. In this context, the translation error of GT reflects the systematic error introduced by the approximation. Moreover, we remove the image-retrieval module from the detection framework. We utilize the confidence scores predicted by the open-vocabulary detector~\cite{liu2023grounding} to identify the detection result from the proposals. As listed in Table~\ref{tab:ana_dete}, this alternative (w/o RGB) leads to more erroneous translation estimations, which demonstrates the effectiveness of our multi-modal object detector.

\begin{table}[!t]
    \normalsize
    \caption{\textbf{Experimental results on the LINEMOD-O~\cite{brachmann2014learning} dataset.} We report the mean angular errors.}
    \centering
    \begin{tabular}{lccc}
    \Xhline{2\arrayrulewidth}
    Method & SuperGlue & 3DAHV & \textbf{DVMNet++} \\
    \hline
    Tran. Error & - & - & 6.47 \\
    Rota. Error & 73.72 & 51.49 & 48.82 \\
    \Xhline{2\arrayrulewidth}
    \end{tabular}
    \label{tab:lmo}
\end{table}

\subsubsection{Robustness to Occlusions}
Given that object pose estimation is often challenged by occlusions, we assess robustness in scenarios involving occlusions by conducting an experiment on the LINEMOD-O~\cite{brachmann2014learning} dataset. The testing data comprises three unseen objects, i.e., cat, driller, and duck. We report the mean angular errors in Table~\ref{tab:lmo}. The results showcase the promising robustness of our method to occlusions.

%% file: sec/5_conclusion.tex
\section{Conclusion}
\label{sec:conclusion}
In this paper, we have introduced DVMNet++, a novel approach for relative pose estimation of unseen objects. Given a single RGB image as the reference, DVMNet++ identifies the object in the query image and computes the relative object pose without relying on the GT object bounding box or rotation hypotheses. This has been achieved via a multi-modal open-set object detector and a deep voxel matching network. Comprehensive experiments on the CO3D, Objaverse, LINEMOD, and LINEMOD-O datasets have demonstrated that our DVMNet++ excels in efficiently delivering accurate relative poses for previously unseen objects.

\noindent\textbf{Acknowledgments.} This work was funded in part by the Swiss National Science Foundation via the Sinergia grant CRSII5-180359 and the Swiss Innovation Agency (Innosuisse) via the BRIDGE Discovery grant 40B2-0\_194729.

%% file: sec/biography.tex
\vspace{-20pt}
\begin{IEEEbiography}
[{\includegraphics[width=1in,height=1.25in,clip,keepaspectratio]{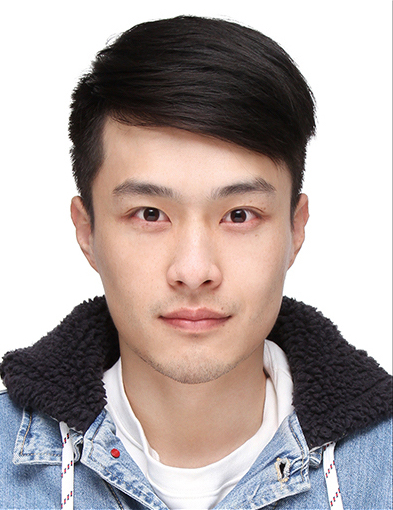}}]{Chen Zhao} is a PhD student at EPFL. He received his BS degree from Huazhong University of Science and Technology in 2017 and his MS degree from Huazhong University of Science and Technology in 2020. His research centered around 3D computer vision, with a specific focus on multi-view geometry and point cloud analysis.
\end{IEEEbiography}

\begin{IEEEbiography}
[{\includegraphics[width=1in,height=1.25in,clip,keepaspectratio]{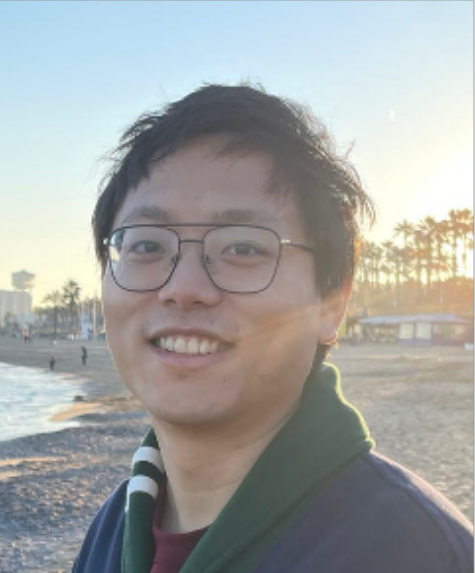}}]{Tong Zhang}  received the B.S. and M.S degree from Beihang University, Beijing, China and New York University, New York, United States in 2011 and 2014 respectively, and he received the Ph.D. degree from the Australian National University, Canberra, Australia in 2020. He is working as a postdoctoral researcher at Image and Visual Representation Lab (IVRL), EPFL. He was awarded the ACCV 2016 Best Student Paper Honorable Mention and the CVPR 2020 Paper Award Nominee. His research interests include subspace clustering, representation learning and 3D vision learning.
\end{IEEEbiography}

\begin{IEEEbiography}
[{\includegraphics[width=1in,height=1.25in,clip,keepaspectratio]{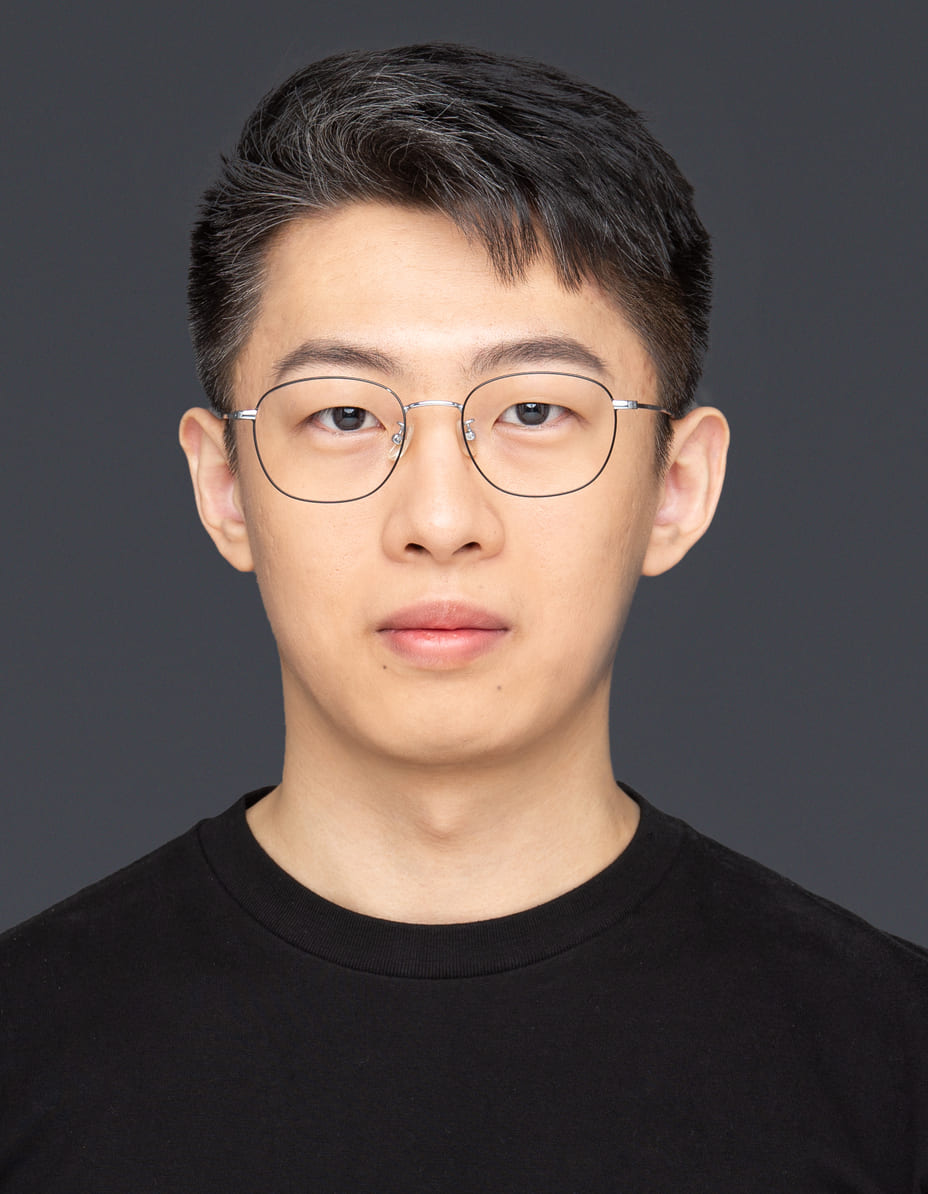}}]{Zheng Dang} is a Postdoctoral Researcher at EPFL. He obtained his B.S. in Automation from Northwestern Polytechnical University in 2014, and his PhD in 2021 from Xian Jiaotong University in Xi’an, China. His research interests lie at the intersection of computer vision, robotics, and deep learning.
\end{IEEEbiography}

\begin{IEEEbiography}[{\includegraphics[width=1in,height=1.25in,clip,keepaspectratio]{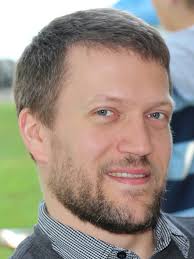}}]{Mathieu Salzmann}is a Senior Researcher at EPFL and an Artificial Intelligence Engineer at ClearSpace. Previously, after obtaining his PhD from EPFL in 2009, he held different positions at NICTA in Australia, TTI-Chicago, and ICSI and EECS at UC Berkeley. His research interests lie at the intersection of machine learning and computer vision.
\end{IEEEbiography}